\documentclass[dvipsnames,format=sigconf,anonymous=false,review=false,nonacm]{acmart}

\AtBeginDocument{%
  \providecommand\BibTeX{{%
    \normalfont B\kern-0.5em{\scshape i\kern-0.25em b}\kern-0.8em\TeX}}}

\copyrightyear{2023}
\acmYear{2023}
\setcopyright{acmlicensed}\acmConference[GECCO '23]{Genetic and
Evolutionary Computation Conference}{July 15--19, 2023}{Lisbon, Portugal}
\acmBooktitle{Genetic and Evolutionary Computation Conference (GECCO '23),
July 15--19, 2023, Lisbon, Portugal}
\acmPrice{15.00}
\acmDOI{10.1145/3583131.3590399}
\acmISBN{979-8-4007-0119-1/23/07}

\usepackage{graphicx}
\usepackage{algorithm,algpseudocode}
\usepackage{subfigure,multirow}
\usepackage[normalem]{ulem}
\useunder{\uline}{\ul}{}
\usepackage{threeparttable}
\def\P{\mathcal{P}}

\newcommand{\rc}[1]{{#1}}

\begin{document}

\title{Local Optima Correlation Assisted Adaptive Operator Selection}

\author{Jiyuan Pei}
% \authornote{}
\email{peijy2020@mail.sustech.edu.cn}
\orcid{0000-0001-9860-5160}
\affiliation{%
  \institution{Southern University of Science and Technology}
  \city{Shenzhen}
  \country{China}
}

\author{Hao Tong}
\email{hxt922@student.bham.ac.uk}
\orcid{0000-0003-4881-8701}
\affiliation{%
  \institution{University of Birmingham}
  \city{Birmingham}
  \country{United Kingdom}
}

\author{Jialin Liu}
% \authornote{}
\email{liujl@sustech.edu.cn}
\orcid{0000-0001-7047-8454}
\affiliation{%
  \institution{Southern University of Science and Technology}
  \city{Shenzhen}
  \country{China}
}

\author{Yi Mei}
\authornote{Corresponding author}
\email{yi.mei@ecs.vuw.ac.nz}
\orcid{0000-0003-0682-1363}
\affiliation{%
  \institution{Victoria University of Wellington}
  \city{Wellington}
  \country{New Zealand}
}

\author{Xin Yao}
% \authornote{}
\email{xiny@sustech.edu.cn}
\orcid{0000-0001-8837-4442}
\affiliation{%
  \institution{Southern University of Science and Technology}
  \city{Shenzhen}
  \country{China}
}

\renewcommand{\shortauthors}{Pei and Tong, et al.}

\begin{abstract}
     For solving combinatorial optimisation problems with metaheuristics, different search operators are applied for sampling new solutions in the neighbourhood of a given solution. It is important to understand the relationship between operators for various purposes, e.g., adaptively deciding when to use which operator to find optimal solutions efficiently. However, it is difficult to theoretically analyse this relationship, especially in the complex solution space of combinatorial optimisation problems. In this paper, we propose to empirically analyse the relationship between operators in terms of the correlation between their local optima and develop a measure for quantifying their relationship. The comprehensive analyses on a wide range of \rc{capacitated vehicle routing problem benchmark instances} show that there is a consistent pattern in the correlation between commonly used operators. Based on this newly proposed local optima correlation metric, we propose a novel approach for adaptively selecting among the operators during the search process. The core intention is to improve search efficiency by preventing wasting computational resources on exploring neighbourhoods where the local optima have already been reached. \rc{Experiments on randomly generated instances and commonly used benchmark datasets are conducted. Results show that the proposed approach outperforms commonly used adaptive operator selection methods.}
\end{abstract}






\maketitle

\section{Introduction}

Metaheuristics in generate-and-test style compose one of the main fields of optimisation. Those algorithms iteratively improve the solutions by replacing the current solution with a new one of better quality. Search operators, like mutation and crossover operators in evolutionary algorithms and differential rules in differential evolution algorithms, contribute the main part in sampling new solutions. Each search operator composes a unique solution neighbourhood, i.e., a subset of the solution space, and the new solutions are sampled from the neighbourhood. As search operators contribute the main part of the search ability of a metaheuristic algorithm, the selection and combination of operators require deliberate design to be efficient. In the literature, to solve unseen problems, several novel operators have been proposed for better handling the unique problem characteristics \cite{croes1958method,pei2021bi}. However, domain knowledge of the specific problem is required, which is inaccessible when facing a new problem. An alternative approach is selecting and utilising existing search operators from relative problems~\cite{thierens2005adaptive,lu2020learning}. Adaptive operator selection (AOS) is a specific research field focusing on dynamically and adaptively selecting search operators during the optimisation process of metaheuristic algorithms with the aim of better utilising the search ability of candidate operators and avoiding resource wasting on unpromising operators. Various AOS approaches have been proposed and proved effective in genetic algorithm \cite{maturana2008compass}, differential evolution \cite{sallam2017landscape,sharma2018performance} and local search \cite{lu2020learning}.

Most existing AOS studies focus on real-valued optimisation problems \cite{fialho2010comparison,sharma2018performance,tan2021differential,zhang2021learning}, while combinatorial optimisation problems, especially routing problems, are seldom considered \cite{lu2020learning,consoli2014diversity}. Because of the continuity, smoothness and numerical representation in real-value optimisation problems, the measurement of operators' effectiveness and similarity of searching states are relatively straightforward \cite{thierens2005adaptive, sallam2017landscape,sharma2018performance}. However, in combinatorial optimisation, the discrete solution space and rugged fitness landscape complicate the operator selection process \cite{consoli2014diversity,lu2020learning}. A recent work \cite{pei2022investigation} suggests that in complex vehicle routing problems, unexpected local optima are reached frequently, leading to the unpromising performance of the commonly used AOS approaches.

Investigation of the operators' neighbourhood, in which the new solutions are sampled, is intuitive for assisting the operator selection. Researches of fitness landscape analysis (FLA) have been  conducted to investigate the characteristics of solution space of combinatorial optimisation problems \cite{tong2022what}. Various approaches have been proposed to quantify the fitness landscape of a given neighbourhood for a given problem \cite{malan2013survey, malan2021survey}, such as local optima network \cite{ochoa2014local}. However, to the best of our knowledge, there is no research focusing on the relationship between different neighbourhoods. Due to the characteristics of search operators, the neighbourhood structures of different operators may own implicit relationships. During optimisation, with the historical record of the exploration by some operators at the early stage, the understanding of the relationship is helpful for operator selection at the later stage.

Taking the capacitated vehicle routing problem (CVRP) as an example, this paper investigates the following three research questions.
\begin{itemize}
    \item In combinatorial optimisation problems, specifically CVRP, does the relationship of the neighbourhood of different search operators exists?
    \item If the relationship exists, how to quantify it?
    \item How to utilise the quantified relationship to enhance metaheuristics?
\end{itemize} 

Local optimum location is one of the key features of a neighbourhood. An efficient optimisation algorithm is expected to detect and avoid local optima to prevent early-maturing convergence. To answer our research questions, this paper focuses on the local optima in each operator's neighbourhood. Specifically, the contributions of this paper are as follows.
\begin{itemize}
    \item A novel measurement of neighbourhood relationship, called \textit{local optima correlation} (LOC), is proposed to analyse and quantify the relationship of search operators.
    \item Empirical analysis of LOC is conducted on a wide range of CVRP benchmark instances. Results indicate that a consistent relationship among a set of commonly used search operators exists.
    \item We also propose a novel LOC-assisted AOS framework. The framework does not rely on specific AOS characteristics, so that most of the existing AOS approaches can be easily adapted into the framework. Experimental study verifies the effectiveness of the proposed framework.
\end{itemize}

The rest of this paper is organised as follows. Section \ref{sec:background} reviews the literature on adaptive operator selection and fitness landscape analysis. The proposed approach and framework are described in Section \ref{sec:methodology}. Section \ref{sec:experiment} presents the experiments and discusses the results. Section~\ref{sec:conclusion} concludes.

\section{Background}\label{sec:background}
This section comprehensively reviews related studies of adaptive operator selection and fitness landscape analysis with a focus on combinatorial optimisation problems, especially CVRP.

\subsection{Adaptive Operator Selection}\label{sec:back:aos}
Adaptive operator selection (AOS) focuses on dynamically and adaptively selecting and applying search operators during the optimisation process so that the computational resource can be assigned to the high-performanced operators at the different searching stages. 
Various AOS approaches have been proposed and can be briefly categorised into two classes \cite{pei2022investigation}, stateless AOS and state-based AOS. 

Stateless AOS approaches record every use of each operator during the optimisation process and estimates the impact of each operator based on the records. Then, the selection probabilities of operators are obtained. There are two main components in stateless AOS approaches: the credit assignment (CA), i.e., how to measure the impact of using an operator, and the operator selection rule (OSR), i.e., how to make a decision based on the credits. Fitness improvement is one of the most classic and commonly used CA, which calculates the credits as the difference of quality between a newly sampled solution and the original solution \cite{thierens2005adaptive}. Several different CA methods have been proposed to handle various problem characteristics based on fitness improvement \cite{fialho2008extreme,soria2014evolvability}. Probability matching (PM) is a widely studied OSR \cite{goldberg1990probability,sharma2018performance, consoli2014diversity}. Its operator selection strategy can be described as Eq. \eqref{equ:pm}, 
\begin{equation}\label{equ:pm}
    \begin{split}
        Q_{i} = & \alpha \cdot r_{i} + (1-\alpha)  Q_{i}, \\
        \P_{i} = & \P_{min} + (1-K\cdot \P_{min})  \frac{Q_{i}}{\sum_{j=1}^K{Q_{j}}},
    \end{split}
\end{equation}
where $K$ is the number of candidate operators, $r_i$ is the reward (i.e., credit) of using operator $i$, and $\P_i$ is the calculated probability of selecting operator $i$. $\P_{min}$ and $\alpha$ are two pre-defined parameters.
Based on PM, adaptive pursuit (AP) is developed for better sensibility \cite{thierens2005adaptive}. AP introduces the parameter $\beta$ to control the sensibility of updating the selection probability $\P_i$, as formulated in Eq. \eqref{equ:ap},
\begin{equation}
\P_{i} = \begin{cases}  \beta \cdot \rc{\P_{max}} + (1-\beta)\P_{i}, &  \text{if } i=\underset{j\in \{1,\dots,K\}}{\arg\max}{Q_{j}},\\
\beta \cdot \P_{min} + (1-\beta)\P_{i}, & \text{otherwise,}
\end{cases}
\label{equ:ap}
\end{equation}
\rc{where $\P_{max}=1-(K-1)\P_{min}$.} With a higher $\beta$, AP reacts more quickly than PM when the performance of operators changes along with the optimisation stage.

In state-based AOS approaches, the features of an optimisation stage (i.e., state) are involved in the decision-making. A mapping from state to decision is learned from training data collected from either the early stage of the current optimisation process or the records of optimising similar problems. The state representation is required to be informative for decision-making and easy to understand by learners. \rc{In state-based AOS approaches for real-valued optimisation \cite{10.1145/3321707.3321813} or assignment problem \cite{handoko2014reinforcement,durgut2021adaptive}, the state representation is straightforward, while in permutation-based combinatorial optimisation (e.g., vehicle routing problems),} extracting state features from sequential solutions in graph-based instances is challenging. \citet{consoli2016dynamic} takes the fitness landscape of the solution population on capacitated arc routing problem (CARP) \cite{zhang2020d,mei2011decomposition} as a state feature to select the crossover operators in evolutionary algorithms. Due to the state directly generated from the whole population, the method of \cite{consoli2016dynamic} is not able to suggest a fine-grained operator for each solution. L2I \cite{lu2020learning}, a deep reinforcement learning approach for operator selection, trains a graph attention network to generate numerical features from CVRP instances, the current solution and the record of previous operator selection, and achieves high performance on randomly generated instances when a large number of instances are accessible for training.

\subsection{Fitness Landscape Analysis}
Fitness landscape analysis (FLA) for combinatorial optimisation problems is a popular approach for understanding the characteristics of a given neighbourhood structure of a solution \cite{malan2013survey, malan2021survey}. It has been successfully applied to several combinatorial optimisation problems \cite{tayaraniN2014landscape}, such as travelling-thief problem \cite{yafrani2018fitness} and dynamic capacitated arc routing problem \cite{tong2022what}.

Many existing FLA approaches focus on sampling a set of representative neighbour solutions (e.g., local optima) of a given solution to investigate the search space. The fitness values of these neighbours are analysed with statistical metrics, such as auto-correlation, fitness distribution, fitness distance correlation, etc., to characterise the fitness landscape \cite{tayaraniN2014landscape}. Recently, a novel FLA approach, called local optima network (LON), has been proposed to consider the fitness landscape as a network \cite{ochoa2014local}. The nodes in the network represent the local optima obtained by local search operators, and the edges indicate the relation between local optima \cite{ochoa2014local}. Therefore, the network analysis and visualisation tools can help people to better understand the problem's landscape.

However, the relationship between different neighbourhoods is seldom investigated. To the best of our knowledge, no existing FLA approach extracts or quantifies the relationship among a neighbourhood set or the corresponding operator set.

\section{Methodology}\label{sec:methodology}

Relationship among operators is instructive for operator selection. With the record of operator performance, the ability of other related operators can be predicted. To fill the gap of research in relationship between operators, and also to assist AOS for better decision-making, we propose a measurement of relationship between search operators, named local optima correlation (LOC). Based on LOC, a novel AOS framework is proposed. 
This section first introduces the local optima correlation (LOC) and then describes the proposed LOC-assisted AOS framework, named AOS-LOC. 

\subsection{Local Optima Correlation}

Given an optimisation problem and a solution, each search operator forms a neighbourhood for this solution, which consists of a set of solutions generated by applying the operator to the given solution.
In a neighbourhood of a specific operator, if the quality of a solution is better than all its neighbours, we say that the solution reaches a local optimum of the operator, and the operator is \textit{local-optimum-trapped} on the solution. For a set of $K$ operators and a solution $s$, we introduce the \textit{local-optimum-trapped vector} $O_s=(o_{s,1},o_{s,2},\dots,o_{s,K})$, $\forall i \in \{1,2,\dots,K\}$, $o_{s,i}\in \{-1,1\}$.  $O_s$ indicates if a local optimum is reached or not for each operator $ope_i$. $o_{s,i}=1$ indicates that $x$ reaches a local optimum in its neighbourhood corresponding to operator $i$. If the local optimum of operator $i$ is not reached, then $o_{s,i}=-1$.

To better represent the characteristics of a solution space, a number of $N$ solutions are sampled from the solution space, forming a solution sequence $S=\{s_1,s_2,\dots,s_N\}$. 
Therefore, a binary matrix $\mathcal{O}$ can be constructed as Eq. \eqref{eq:LOmatrix}.
\begin{equation} \label{eq:LOmatrix}
\mathcal{O} = \begin{bmatrix} 
    O_{s_1}\\ 
    \vdots \\ 
    O_{s_N}
    \end{bmatrix}
    =
    \begin{bmatrix} 
    o_{s_1,1} & \cdots & o_{s_1,K}\\ 
    \vdots  & \ddots &\vdots\\ 
    o_{s_N,1} & \cdots &  o_{s_N,K}
    \end{bmatrix}.
\end{equation}
Each column $C_i=[o_{s_1,i},o_{s_2,i},\dots,o_{s_N,i}]^T$ of $\mathcal{O}$ indicates if the local optimum is reached or not for operator $i$ on each sampled solution $s_j$, $\forall j \in \{1,2,\dots,N\}$.

Taking $C_i$ and $C_j$ as two binary sequences for any operators $i$ and $j$, the \textit{local optima correlation (LOC)} can be calculated by Eq. \eqref{equ:correlation}.
\begin{equation}\label{equ:correlation}
    \mathcal{LOC}_{i,j} = corr(C_i,C_j)=\frac{1}{|S|}\sum_{s_k\in S}o_{s_k,i}o_{s_k,j}.
\end{equation}
$\mathcal{LOC}_{i,j} = 1$ indicates that the local optima of operators $i$ and $j$ are always reached simultaneously, i.e., $C_i$ and $C_j$ are perfectly positively correlated. $\mathcal{LOC}_{i,j} = -1$ indicates that when any of the operators $i$ or $j$ is trapped by a local optimum, the other one is always able to find a better neighbour solution, i.e., $C_i$ and $C_j$ are perfectly negatively correlated. We form $\mathcal{LOC}_{i,j}$ for all operators into the local optima correlation matrix $\mathcal{LOC}$ as Eq. \eqref{eq:coormatrix}. 
\begin{equation} \label{eq:coormatrix}
\mathcal{LOC} = \begin{bmatrix} 
    \mathcal{LOC}_{1,1} & \cdots & \mathcal{LOC}_{1,K}\\ 
    \vdots  & \ddots &\vdots\\ 
    \mathcal{LOC}_{K,1} & \cdots & \mathcal{LOC}_{K,K}
    \end{bmatrix}.
\end{equation}
$\mathcal{LOC}$ is an upper triangular matrix since $corr(X,Y)=corr(Y,X)$ and the main diagonal values are $1$ as $corr(X,X)=1$.

For a sampled solution set, $\mathcal{LOC}$  reflects the implicit relationship between operators on those solutions. When the LOC matrices perform high similarity between different sampling, we conclude that the relationship found is consistent on the problem. Furthermore, if the LOC matrices follow the same pattern between different problems, we conclude that the relationship found is possibly general and doesn't highly reply on the problem itself.

\subsection{LOC-assisted Adaptive Operator Selection}

\begin{algorithm}
\caption{General framework of local search with AOS.\label{algo:LS-AOS}}
\begin{algorithmic}[1]
\Require{a set of $K$ search operators $OPE = \{ope_1,\dots,ope_K\}$, a problem and its solution evaluator $fitness()$, iteration budget $max\_ite$, a base AOS approach which can generate selection probability by AOS.decision\_making() and update record by AOS.record\_update()}
\State $s_0 \leftarrow$ solution initialisation
\For{$ite \leftarrow 1$ to  $max\_ite$} % iterate T step
    \State $\mathcal{P} \leftarrow $ AOS.decision\_making() \label{line:LS-AOS_aos_OSR}
    \State $ope \leftarrow $randomly\_select$(OPE,\mathcal{P})$  \label{line:LS-AOS_prob_select}
    \State $s_{ite} \leftarrow ope(s_{ite-1})$
    \State $r_{ite} \leftarrow fitness(s_{ite}) - fitness(s_{ite-1})$ \label{line:LS-AOS_FI}
    \State AOS.record\_update$(ope,r_{ite})$ 
\EndFor
\State \Return $s_{max\_ite}$
\end{algorithmic}
\end{algorithm}

Algorithm \ref{algo:LS-AOS} demonstrates the classic framework for applying AOS in metaheuristics, taking local search as an example. At the beginning of each optimisation iteration, the selection probability of each operator is calculated based on the historical records (line \ref{line:LS-AOS_aos_OSR}). An operator is selected according to the probability (line \ref{line:LS-AOS_prob_select}). Then a new solution is generated by applying the selected operator to the current solution and evaluated. The performance of the selected operator is evaluated by CA (cf. Section \ref{sec:back:aos}). In Algorithm \ref{algo:LS-AOS}, the fitness improvement is used (line \ref{line:LS-AOS_FI}), but other evaluation measures can be adopted. At the end of this iteration, the operator selection and performance are used to update the historical records in AOS.

The work of \cite{pei2022investigation} investigates the behaviour and characteristic of AOS approaches and suggests that in combinatorial optimisation problems, especially complex vehicle routing problems, there are usually many local optima that are reached unexpectedly, leading to the failure of operator performance estimation. Specifically, the performance of an operator with high historical performance may rapidly drop to zero as a consequence of reaching local optima. However, AOS approaches will still assign a high selection probability to such an operator for a relatively long period due to its high performance in history. Another operator will be assigned a high selection probability only after the reputation of the local-optimum-trapped operator decreases due to the considerable resource wasted on it. In this case, the performance of AOS approaches diminishes.

\begin{algorithm}[htbp]
\caption{General framework of local search with LOC-assisted AOS.\label{algo:LS-LOCAOS}}
\begin{algorithmic}[1]
\Require{a set of $K$ search operators $OPE = \{ope_1,\dots,ope_K\}$,  a problem instance and an evaluator $fitness()$, iteration budget $max\_ite$, a base AOS approach which can generate selection probability by AOS.decision\_making() and update record by AOS.record\_update(), matrix $\mathcal{LOC}$ }
\State $LO \leftarrow \emptyset $
\State $s_0 \leftarrow$ solution initialisation
\For{$ite \leftarrow 1$ to  $max\_ite$} % iterate T step
    \State $P \leftarrow $ AOS.decision\_making() 
    \For{$i \in LO$}
        \State $\P \leftarrow \P \times \rc{[1-\mathcal{LOC}_{i,1},\dots,1-\mathcal{LOC}_{i,K}]}$ \label{line:LS-AOS-LOC_times_LOC}
    \EndFor
    \State $\P \leftarrow \P/(sum(\P) $
    \State $ope \leftarrow$ randomly\_select$(OPE,\P)$
    \State $s_{ite} \leftarrow ope(s_{ite-1})$
    \State $r_{ite} \leftarrow fitness(s_{ite}) - fitness(s_{ite-1})$
    \State AOS.record\_update$(ope,r_{ite})$
    \If {$r_{ite} > 0$} \label{line:LS-AOS-LOC_LO_update_begin}
        \State $LO \leftarrow \emptyset$
    \Else 
        \State $LO \leftarrow LO + ope$
    \EndIf \label{line:LS-AOS-LOC_LO_update_end}
\EndFor
\State \Return $s_{max\_ite}$
\end{algorithmic}
\end{algorithm}

For handling the aforementioned issue, predicting local-optimum-trapped operators based on the previous records and avoiding exploring trapped operators is an intuitive strategy. If an operator $ope_i$ is confirmed as local-optimum-trapped after trying it on the current solution, other operators $ope_j$ that own high LOC (cf. Eq. \eqref{equ:correlation}) with $ope_i$ are also likely to be trapped on the current solution. We refer to those operators as \textit{likely trapped operators}. Therefore, we propose to assist the operator selection by LOC, as demonstrated in Algorithm \ref{algo:LS-LOCAOS}. An operator set $LO$ is maintained to store the operators with which the current solution has reached a local optimum. The selection probability calculated by AOS is modified by multiplying vector $\rc{[1-\mathcal{LOC}_{i,1},\dots,1-\mathcal{LOC}_{i,K}]}$ for each operator $i$ in $LO$ (line \ref{line:LS-AOS-LOC_times_LOC}). For any operators $i$ and $j$, the range of $(1-\mathcal{LOC}_{i,j})$ is $[0,2]$. The lower the correlation of local optima of $i$ and $j$ is, the higher $(1-\mathcal{LOC}_{i,j})$ is. Then, the probabilities of selecting likely trapped operators decrease and the probabilities of selecting other operators increase. In this way, less resource will be wasted. After applying an operator, $LO$ will be updated, as in lines \ref{line:LS-AOS-LOC_LO_update_begin}-\ref{line:LS-AOS-LOC_LO_update_end}.

In the proposed LOC-assisted AOS framework, the coupling of AOS and LOC is considerably low, so most AOS approaches that output selection probability can be easily embedded into our proposed framework. It gives the framework high generality and can be applied to various problems and AOS approaches.

\section{Experimental Studies}\label{sec:experiment}

Values in the LOC matrix of a given problem depend on the solution sampling. Operators' performance on different solution sets are different, which is a major factor that affects the stability and generalisation of LOC. Hence, a stable LOC that is insensitive to solution sampling is expected. We take capacitated vehicle routing problem (CVRP), one of the most classic combinatorial optimisation problems, as a test case, and conduct experiments to investigate the characteristics of LOC matrix on a specific operator set. Then, the obtained LOC is adopted into the proposed LOC-assisted AOS framework to verify the optimisation ability of the framework. 
This section first presents the experiment setting, and then presents the empirical analysis of LOC on a wide range of CVRP benchmark instances. Finally, the performance of LOC-assisted AOS framework is presented and discussed.

\subsection{Experiment Setting}\label{sec:setting}

\begin{table*}[ht]
    \centering\setlength{\tabcolsep}{20pt}
    \centering
    \caption{\label{tab:operators}Candidate operators for CVRP\cite{lu2020learning}.}
    \begin{tabular}{ccl}
\toprule
\textbf{Index} & \textbf{Operator(\#operated routes)}                   & \multicolumn{1}{c}{\textbf{Description}}                                                                              \\ \midrule
1     & 2opt                  & Reverse a section of a given route                                                       \\ 
2     & Symmetric-exchange(1)  & Exchange two   customers in a given route                                                         \\ 
3     & Relocate(1)            & Move one customer to another location in the given route                                       \\ 
4     & Cross/Reverse-cross(2) & Exchange the end   customers of two given routes                                               \\ 
5-7   & Symmetric-exchange(2)  & Exchange sections with same length in 1, 2, 3   between two routes                                    \\ 
8-10  & Relocate(2)            & Move a section with length 1, 2, 3  from a given route to another one            \\ 
11    & Cyclic-exchange(3)     & Exchange  customers between three given routes                                                          \\ 
12-17 & Asymmetric-exchange(2) & Exchange sections with different length in 1, 2, 3 between two routes%, the two sections have the different length 
\\ \bottomrule
\end{tabular}
\end{table*}

\rc{We design two experiments, (i) empirical study of LOC's pattern, to verify the stability and capability of LOC to extract operators' characteristics among different problem instances, and (ii) testing of AOS-LOC framework, to verify the effectiveness of LOC to assist the decision making of AOS.}

\subsubsection{Problem instances} In this paper, Euclidean CVRPs are considered as the test problem. A CVRP instance can be formed into a graph $G=(V,E)$ with a set of capacitated vehicles. Each vertex in $V$, called customer, owns a numerical feature named demand. Each edge in $V$ represents a road connecting two customers and owns a travel cost. The objective is to find a routing plan that minimises the total travel cost of all vehicles, without violating the capacity constraint of any vehicle. Euclidean CVRP is a specific subset of CVRPs in which any two vertices are connected by an edge and the travel cost is the Euclidean distance between the two vertices. Readers are referred to \cite{toth2002models} for the mathematical model of CVRP. Various metaheuristics with multiple operators have been proposed to solve CVRPs \cite{lu2020learning,consoli2014diversity}, making it a good case for investigating AOS and metaheuristics in solving combinatorial optimisation problems. 
The uniformly randomly generated CVRP instances as in \cite{lu2020learning}, referred to as UniRand, together with the commonly used benchmark Li instances \cite{li2005very}, Loggi and ORTEC instances\footnote{\label{first_footnote}Loggi and ORTEC instances: http://dimacs.rutgers.edu/programs /challenge/vrp/cvrp/} are used for studying LOC and LOC-assisted AOS, as listed in Table \ref{tab:instances}. \rc{For both the solution sampling in the first experiment and optimisation in the second experiment, processes stop at the 40,000th iteration on UniRand instances, as suggested in \cite{lu2020learning}. Processes stop at the 2000th iteration on Li, Loggi and ORTEC instances due to the significantly larger instance size.}

\begin{table}[htbp]
    \centering\setlength{\tabcolsep}{2pt}
    \centering
    \caption{CVRP benchmark instances \cite{li2005very}.\label{tab:instances}}
    \begin{tabular}{cccccc}
\toprule
\multirow{2}{*}{\textbf{Instance}} & \multirow{2}{*}{$\mathbf{|V|}$} & \textbf{Minimum} & \multicolumn{1}{c}{\multirow{2}{*}{\textbf{{Instance}}}} & \multirow{2}{*}{$\mathbf{|V|}$} & \textbf{Minimum} \\
&   & \textbf{\#vehicles} & &   & \textbf{\#vehicles} \\
\midrule
Li-21           & 560   & 10            &
Loggi-n401-k23   & 400   & 23           \\
Li-22           & 600   & 15             &
Loggi-n501-k24   & 500   & 24             \\
Li-23           & 640   & 10             &
Loggi-n601-k19   & 600   & 19             \\
Li-24           & 720   & 10             &
Loggi-n601-k42   & 600   & 42             \\
Li-25           & 760   & 19             &
Loggi-n901-k42   & 900   & 42            \\
Li-26           & 800   & 10             &
Loggi-n1001-k31  & 1000  & 31             \\
Li-27           & 840   & 20             &
ORTEC-n242-k12   & 241   & 12             \\
Li-28           & 880   & 10             &
ORTEC-n323-k21   & 322   & 21             \\
Li-29           & 960   & 10             &
ORTEC-n405-k18   & 404   & 18             \\
Li-30           & 1040  & 10             &
ORTEC-n455-k41   & 454   & 41             \\
Li-31           & 1120  & 10             &
ORTEC-n510-k23  & 509   & 23             \\
Li-32           & 1200  & 11             &
ORTEC-n701-k64  & 700  & 64             \\
\bottomrule
\end{tabular}
\end{table}

\subsubsection{Search operators} The comprehensive set of search operators for AOS in CVRPs provided by \citet{lu2020learning} (Table \ref{tab:operators}) are used in this paper as it is the state-of-the-art work in AOS for CVRPs. We take the local search algorithm and operators in \cite{lu2020learning} as a case study of LOC and LOC-assisted AOS. In our experiments, all setting of the algorithm are the same as in \cite{lu2020learning} unless otherwise stated.

\subsubsection{Solution sampling for LOC} The solution set used to generate LOC matrix does not require a specific sampling method. \rc{In the first experiment}, the solution set is obtained by recording solutions during the optimisation process of the algorithm proposed in \cite{lu2020learning} with one modification, thus all operators are applied on each solution for sampling, while in the original algorithm, only one single operator is selected and applied. 
As operators will travel the neighbourhood to find better solutions, $O_{s_j}$ (matrix element in Eq. \eqref{eq:LOmatrix}) is obtained and one of the found neighbours of higher quality will be randomly selected, denoted as $s_{j+1}$, and replaces the current solution $s_j$. Solutions are recorded by repeating the process. \rc{If all operators are local-optima trapped on a specific solution, the solution will not be recorded for calculating LOC, as it does not provide essential information for analysis.} Then, a LOC matrix can be calculated based on the selected set. For a single instance, the LOC calculation requires more computational resources than the optimisation process for solving the instance. \rc{However, if LOC matrices can extract general features between different instances and perform similar patterns, the LOC matrices can be used inter-instance to assist AOS, i.e., calculate LOC from seen small instances and then use the LOC matrix in unseen, larger instances. Therefore in the first experiment,} the LOC matrix from different sampling on the same instance and different sampling on different instances are calculated and compared.

\subsubsection{Validation of AOS-LOC} \rc{In the second experiment,} the proposed AOS-LOC is compared to base AOS approaches. L2I, the improvement operator selector proposed in \cite{lu2020learning}, and two commonly used stateless AOS approaches, PM and AP, are implemented as base approaches. Notable L2I is trained and tested only on UniRand instances\rc{, since it requires a large number of training instances following the same distribution as the target instance}. AOS-LOCs with PM, AP and L2I as the base approach are named as PM-LOC, AP-LOC and L2I-LOC, respectively. \rc{Base AOS approaches in AOS-LOC framework share the same parameter setting as they are used along.} Parameters in PM and AP are arbitrarily set as $\P_{min} = \frac{0.5}{K-1}$, $\alpha = 0.2$ and $\beta = 0.2$. 
%\rc{with the reason that LOC is the focus of this study instead of base AOS approaches and in real-world problem-solving scenarios the resource may not be enough for parameter tuning}. 
L2I is trained and tested with the code\footnote{\rc{Code of L2I is provided by \cite{lu2020learning}, available at github.com/rlopt/l2i}} provided by \cite{lu2020learning} and the parameters are set as suggested in \cite{lu2020learning}.

\subsection{LOC Pattern}

On all benchmark instances introduced in Section \ref{sec:setting}, ten independent solution sampling trials are conducted and the corresponding LOC matrix for each sampled solution set is calculated. 
Kendall correlation coefficient is commonly used for evaluating the ranking similarity of two paired data. It is applied to a pair of LOC matrices to evaluate their similarity. Specifically, for each operator, the Kendall correlation coefficient of corresponding rows in two LOC matrices is calculated. Then, the average value of all operators is calculated as the similarity of the two LOC matrices. The mean and variance of similarity of each pair of sampling are shown in Table \ref{tab:samplingKendalltauSimilarity}. The LOC from different sampling for an instance is consistent.

\begin{table}[htbp]
    \centering\setlength{\tabcolsep}{2pt}
    \centering
    \caption{Pairwise similarity of ten LOC matrices on each instance. LOC matrices from different solution sampling are highly consistent.\label{tab:samplingKendalltauSimilarity}}
    \begin{tabular}{cccrcc}
 \toprule
\textbf{Instance}        & \textbf{Mean}  & \textbf{Variance} & \multicolumn{1}{c}{\textbf{Instance}}        & \textbf{Mean}  & \textbf{Variance}      \\\midrule
Li-21           & 0.845 & $\approx 0$ & 
Loggi-n401-k23  & 0.864 & $\approx 0$ \\
Li-22           & 0.851 & 0.002       & 
Loggi-n501-k24  & 0.857 & $\approx 0$ \\
Li-23           & 0.840  & 0.001       & 
Loggi-n601-k19  & 0.842 & 0.001       \\
Li-24           & 0.853 & $\approx 0$ & 
Loggi-n601-k42  & 0.843 & $\approx 0$ \\
Li-25           & 0.859 & 0.001       & 
Loggi-n901-k42  & 0.833 & 0.002       \\
Li-26           & 0.863 & $\approx 0$ & 
Loggi-n1001-k31 & 0.841 & 0.001      \\
Li-27           & 0.862 & $\approx 0$ & 
ORTEC-n242-k12  & 0.872 & $\approx 0$ \\
Li-28           & 0.837 & $\approx 0$ & 
ORTEC-n323-k21  & 0.865 & $\approx 0$ \\
Li-29           & 0.861 & $\approx 0$ &
ORTEC-n405-k18  & 0.852 & $\approx 0$ \\
Li-30           & 0.865 & $\approx 0$ &
ORTEC-n455-k41  & 0.830  & 0.002       \\
Li-31           & 0.830  & $\approx 0$ &
ORTEC-n510-k23  & 0.866 & $\approx 0$ \\
Li-32           & 0.829 & $\approx 0$ &
ORTEC-n701-k64  & 0.793 & 0.003       \\
&&&UniRand & 0.880 & 0.002 \\
\bottomrule
\end{tabular}
\end{table}

\begin{figure*}[htbp]
\centering
\begin{minipage}[t]{0.33\textwidth}
\centering
\includegraphics[width = \textwidth]{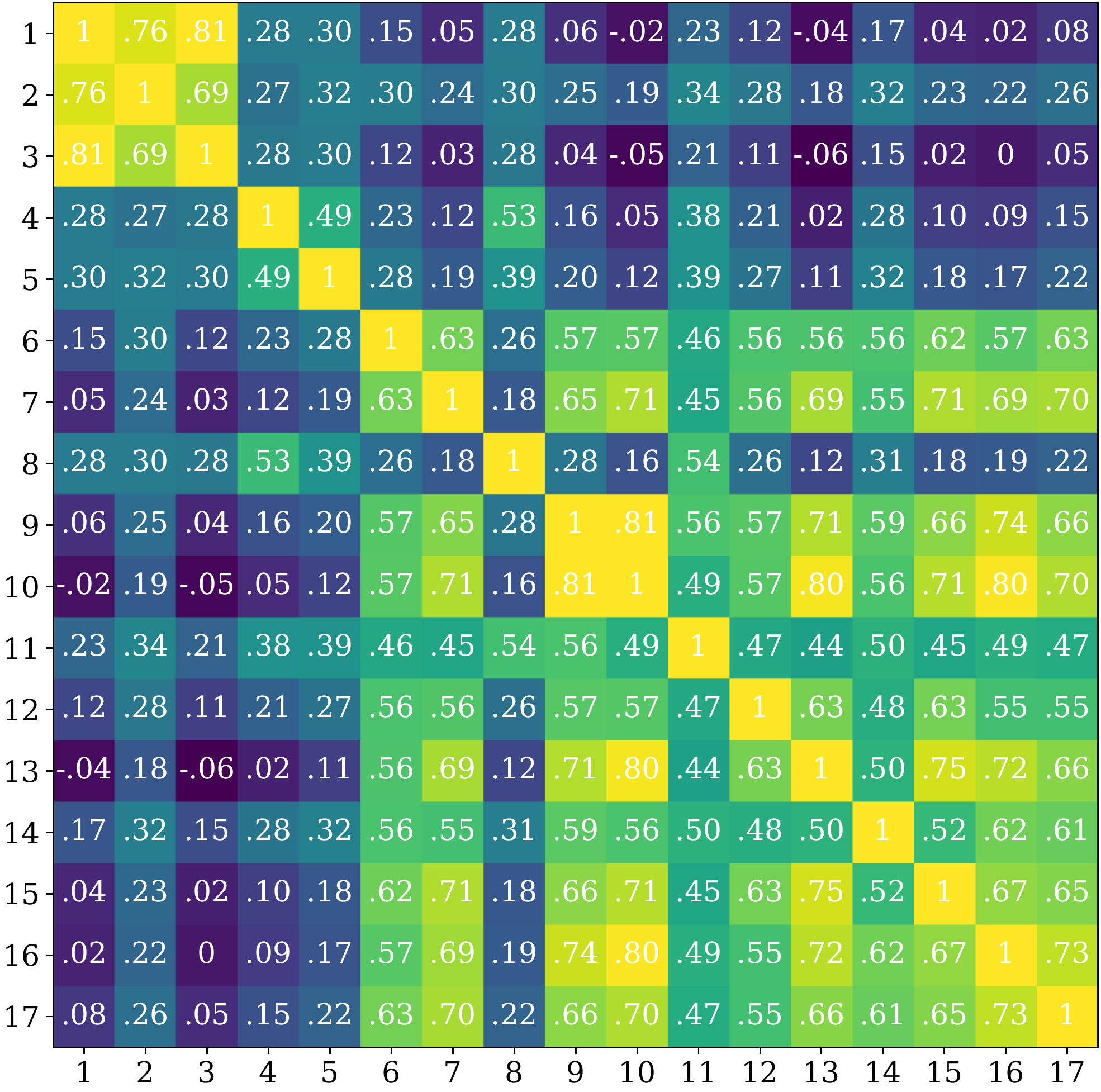}
\end{minipage}\hfill
\begin{minipage}[t]{0.33\textwidth}
\centering
\includegraphics[width = \textwidth]{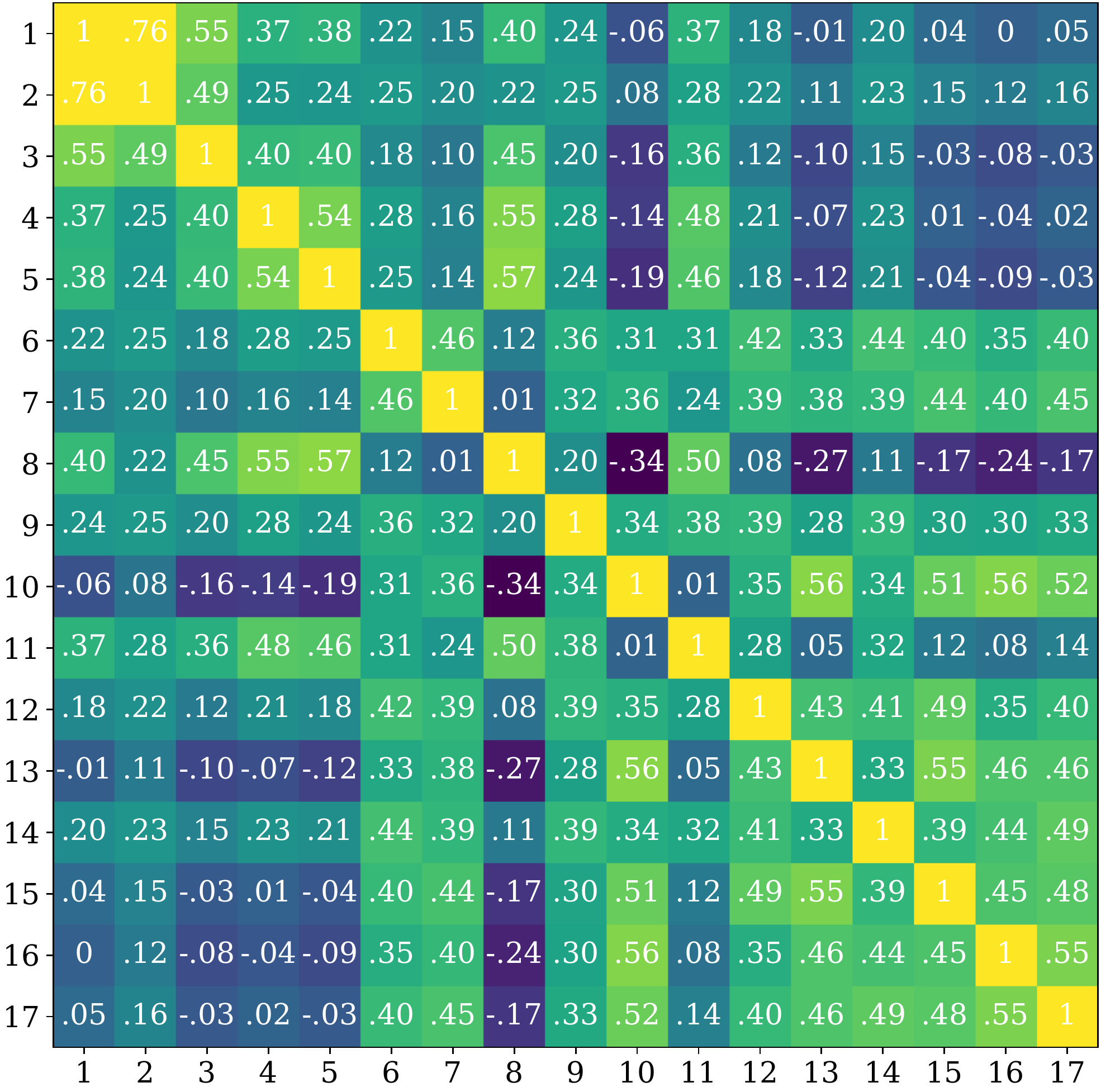}
\end{minipage}\hfill
\begin{minipage}[t]{0.33\textwidth}
\centering
\includegraphics[width = \textwidth]{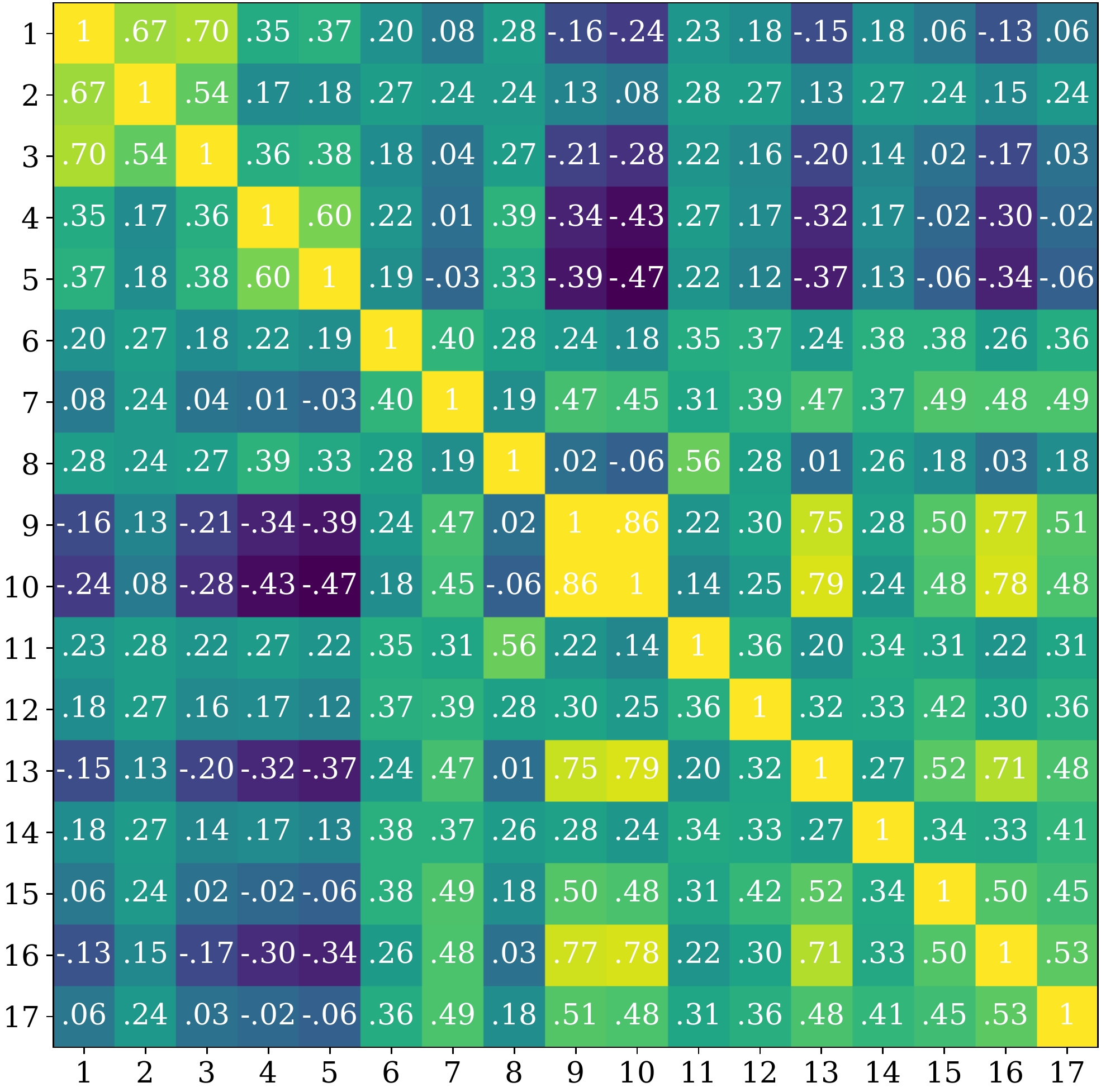}
\end{minipage}
\caption{Illustrative examples of average LOCs from three instances namely UniRand, Li-27 and ORTEC-n701-k64 from left to right. Though the values are different, similar patterns are observed in the rank of rows.\label{fig:avgLOCs}}
\end{figure*}

To compare the similarity of LOC matrices from different instances, the average LOC matrix of the ten samplings is calculated for each instance. As illustrative examples, Fig. \ref{fig:avgLOCs} shows the average LOC matrices of three instances. Though the values in LOC matrices are different, the rank of items in each row shares a similar pattern. For example, the three single-route-operators, \textit{2opt}, \textit{symmetric-exchange} and \textit{relocate} indexed with 1, 2 and 3, respectively, have a high correlation to each other. The operators 4 and 5 (i.e., \textit{cross/reverse-cross} and \textit{symmetric-exchange} with section length 1) also always have a high correlation. The operator indexed with 8, \textit{relocate} with section length 1, always has a different pairwise correlation compared to operators 7 and 9 (i.e., \textit{symmetric-exchange} with section length 3 and \textit{relocate} with section length 2), though they are the same operation with different parameters. It verifies that the relationship among operators has an implicit pattern.

\begin{figure}[ht]
\includegraphics[width = 1\linewidth]{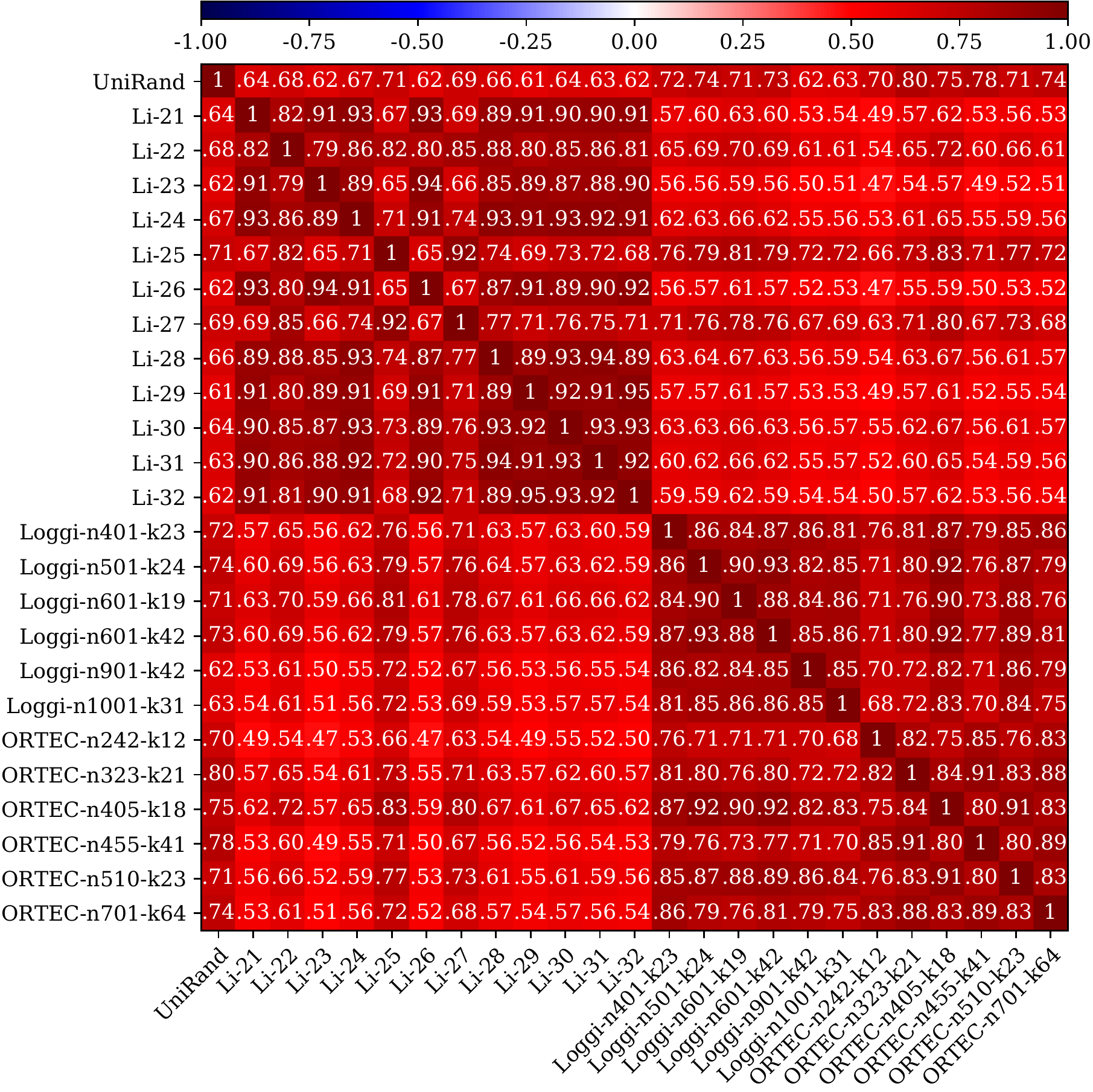}
\caption{Average similarity of LOC matrices on each instance pair. Instances belong to the same group have obviously higher LOC similarity. Notable the range of similarity is [-1,1], where ``-1'' indicates completely opposite and ``1'' indicates identical. The minimum value of this matrix is 0.47.\label{fig:instanceKendalltauSimilarity}}
\end{figure}

Fig. \ref{fig:instanceKendalltauSimilarity} summarises the similarity of each pair of the average LOC matrix. 
For any two instances, the similarity is high than 0.46, representing relatively high relevancy. The instances from the same benchmark have a higher similarity. It indicates that the obtained LOC matrices obtain consistent information which is independent of the problem instance. We conclude that the proposed LOC metrix can represent the universal relationship of a given operator set.

\subsection{Performance of LOC-assisted AOS}

AOS-LOC framework works to solve a problem instance with a given LOC matrix. \rc{Previous experiment has proved that LOC is capable of extracting instance-independent features of operators' behaviour.} 
%Since commonly the target problem instance is unaccessible for sampling and obtaining LOC matrix \rc{and randomly generated instances are always available} 
\rc{Since randomly generated instances are always available for analysis while similar-characteristic instances are not}, in the following experiments we take a LOC matrix calculated from one single sampling from a UniRand instance for all implementation of AOS-LOC \rc{on the UniRand, Li, Loggi and ORTEC instances. }

\begin{figure*}[htbp]
\includegraphics[width = 0.7\linewidth]{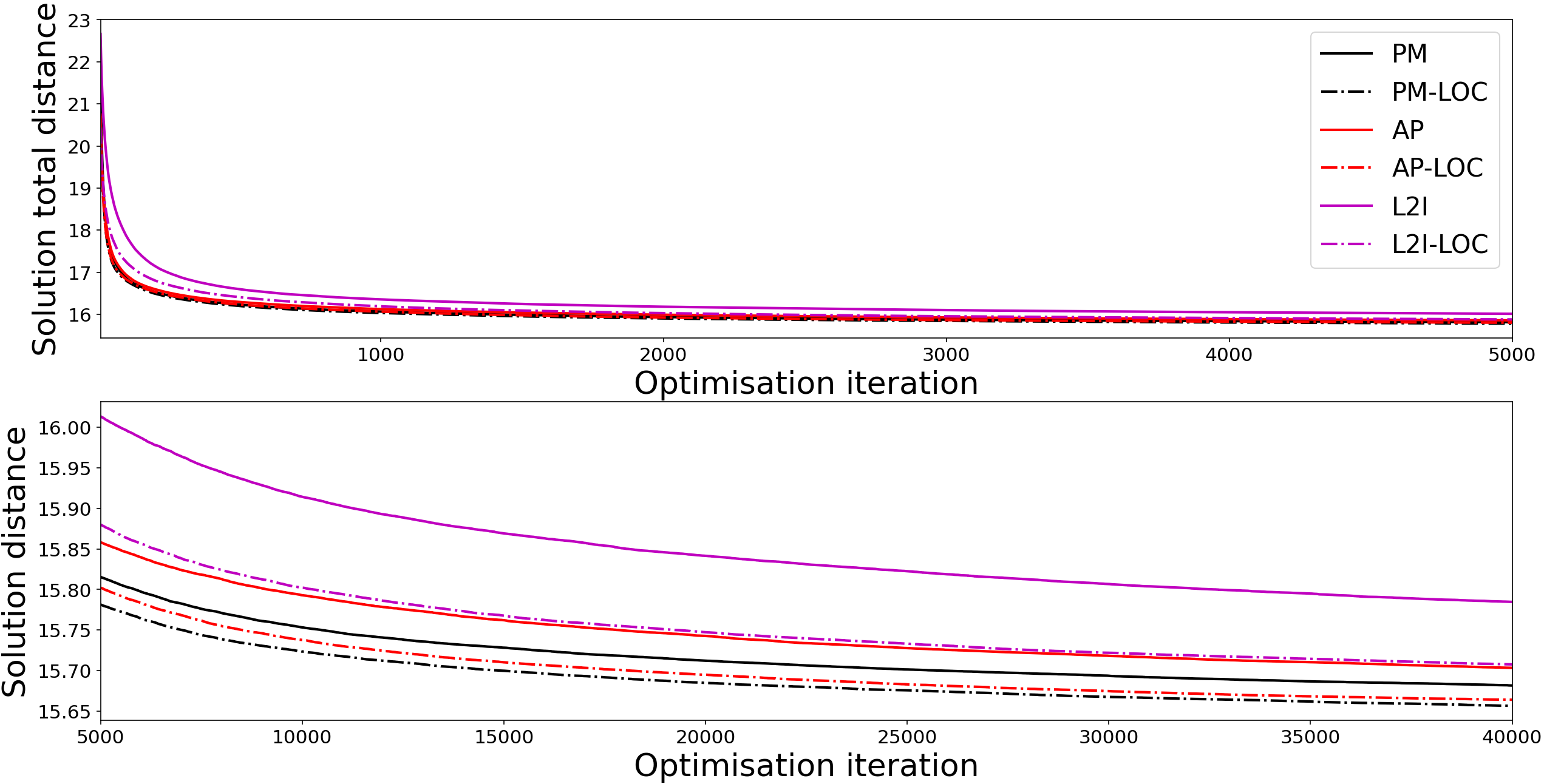}
\caption{Convergence curves of AOS approaches averaged over 2000 generated UniRand CVRP-100 instances (top: 0-5000 iterations; bottom: 5000-40000 iterations). LOC (dashed curves) significantly improves all the base AOS approaches (solid curves). \label{fig:unirandConvergence}}
\end{figure*}

2000 UniRand instances are generated for testing. PM, AP, L2I, PM-LOC, AP-LOC and L2I-LOC are applied to solve each instance once. Fig. \ref{fig:unirandConvergence} demonstrates the average convergence over the 2000 instances. By applying the proposed AOS-LOC framework, the performances of all the three base AOS approaches are improved. Wilcoxon signed-rank test indicates that the quality of final output solution of PM-LOC, AP-LOC and L2I-LOC is significantly better than the ones of PM, AP and L2I respectively, with $p<0.05$.

\begin{table}[htbp]
    \centering
    \caption{Average number of times that local-optimum-trapped operators are selected during optimisation. LOC improves the base AOS approaches on all instances in terms of this metric. The difference between each base AOS and the corresponding AOS-LOC is significant under Wilcoxon signed-rank test with  $p<0.05$. \label{tab:num0reward}}
    \begin{tabular}{ccccc}
 \toprule
\textbf{Instance}        &  \textbf{AP} &  \textbf{AP-LOC} &  \textbf{PM} &  \textbf{PM-LOC} \\\midrule
Li-21           & 886.4                  & 865.3                      & 921.7                  & 888.4                      \\
Li-22           & 934.6                  & 931.8                      & 973.3                  & 967                        \\
Li-23           & 867.9                  & 884.7                      & 913.5                  & 893                        \\
Li-24           & 940.8                  & 928.5                      & 968.7                  & 958.5                      \\
Li-25           & 989.5                  & 986.6                      & 1022                   & 1005.9                     \\
Li-26           & 942.9                  & 913.6                      & 955.5                  & 950.1                      \\
Li-27           & 1025.5                 & 993.6                      & 1037.3                 & 1029.5                     \\
Li-28           & 994.8                  & 953.3                      & 1021.9                 & 1005.7                     \\
Li-29           & 989.6                  & 969                        & 1011.9                 & 999.5                      \\
Li-30           & 1034.8                 & 1016.2                     & 1059.9                 & 1015.4                     \\
Li-31           & 1015.6                 & 987.3                      & 1053.7                 & 1020.5                     \\
Li-32           & 1012.3                 & 983.2                      & 1046.4                 & 997.9                      \\
Loggi-n401-k23  & 885.7                  & 876.4                      & 929.9                  & 910.7                      \\
Loggi-n501-k24  & 914.7                  & 904.9                      & 963.9                  & 936.7                      \\
Loggi-n601-k19  & 965.1                  & 940.6                      & 994.8                  & 980.5                      \\
Loggi-n601-k42  & 882.7                  &  {875.7}  & 908.1                  &  {913.2}  \\
Loggi-n901-k42  & 1022.9                 &  {991.3}  & 1041                   &  {1026.3} \\
Loggi-n1001-k31 & 1035.9                 &  {1017.2} & 1084.9                 &  {1052.8}\\
ORTEC-n242-k12  & 820.3                  & 804.6                      & 861.7                  & 837.5                      \\
ORTEC-n323-k21  & 772                    & 765.8                      & 800.6                  & 801.3                      \\
ORTEC-n405-k18  & 846.1                  & 840.9                      & 896.8                  & 878.1                      \\
ORTEC-n455-k41  & 783.2                  & 766.5                      & 795.1                  & 800.4                      \\
ORTEC-n510-k23  & 897.2                  & 881.1                      & 935.9                  & 914.9                      \\
ORTEC-n701-k64  & 846.5                  &  {844.5}  & 879.2                  &  {867.5}  \\

\bottomrule
\end{tabular}
\end{table}

AOS approaches are also tested on benchmark instances, except for L2I as it requires training instances that follow the same probability distribution as test instances. Each approach is tested independently for 30 repeats on each instance. The 30 final output solution distances of each approach are tested by Wilcoxon signed-rank test. Table \ref{tab:benchmarkStatisticsTest} summarises the comparison results of each AOS and AOS-LOC implementation in terms of the average final distance of the 30 repeat trials. On all the instances, AOS-LOC approaches achieve better performance than the corresponding base AOS approaches. To the phenomenon that on multiple instances the difference is not significant enough by statistics test, a possible reason is the huge uncertainty due to both the random factor in operators and the probabilistic selection. Besides, the differences between operators also affect. A diverse operator set may enlarge the performance difference between selection approaches.

\begin{table}[htbp]
    \centering
    \caption{Average number of selecting a local-optimum-trapped operator during optimisation on 2000 generated UniRand CVRP-100 instances. The difference of each AOS-LOC and corresponding base AOS is significant under Wilcoxon signed-rank test with  $p<0.05$.\label{tab:num0reward1}}
 \begin{tabular}{cccc}
 \toprule
\textbf{AOS} & \textbf{Avg. \#trapped} & \textbf{AOS} & \textbf{Avg. \#trapped} \\\midrule
AP                      & 3313.4 & AP-LOC                  & 2976.3 \\
PM                      & 3649.1 & PM-LOC                  & 3525.6       \\
L2I                     & 3457.4 & L2I-LOC                 & 3274.6\\\bottomrule
\end{tabular}
\end{table}

As introduced above, LOC only affects the selection when at least one local-optimum-trapped operator is selected and tried on a given solution. Therefore, we summarise the number of times that trapped operators are selected after at least one other trapped operator is tried, as listed in Tables \ref{tab:num0reward} and \ref{tab:num0reward1}. The core intention of LOC in AOS-LOC is to predict local optima and avoid selecting local-optimum-trapped operators, the number is expected to be lower than the one of base AOS. Experimental results verify the effectiveness of LOC in AOS-LOC.

\begin{table}[htbp]
    \centering
    \caption{Averaged distances of final solutions found using different AOS approaches over 30 trials. Bold/underlined indicates that the AOS-LOC is better/worse than the corresponding AOS approach, respectively. ``$+$'' and ``$-$'' highlight the cases where the difference is significant under the Wilcoxon signed-rank test with $p<0.05$.\label{tab:benchmarkStatisticsTest}}
    \begin{tabular}{crlrl}
     \toprule
\textbf{Instance}        & \multicolumn{1}{c}{\textbf{AP}}       & \multicolumn{1}{l}{\textbf{AP-LOC}}                 & \multicolumn{1}{c}{\textbf{PM}}       & \multicolumn{1}{l}{\textbf{PM-LOC}}                \\\midrule
Li-21           & 16875 & \textbf{16757} & 16740 & \textbf{16715} \\
Li-22           & 14432 & \textbf{14412} & 14392 & \textbf{14387} \\
Li-23           & 19504 & \textbf{19461} & 19442 & \textbf{19434} \\
Li-24           & 22651 & \textbf{22434} (+)      & 22476 & \textbf{22423} \\
Li-25           & 16433 & \textbf{16391} & 16366 & \textbf{16366} \\
Li-26           & 25368 & \textbf{25028} (+)      & 24944 & {\ul 24977}    \\
Li-27           & 17179 & \textbf{17123} & 17105 & {\ul 17119}    \\
Li-28           & 28683 & \textbf{28449} (+)      & 28321 & \textbf{28252} \\
Li-29           & 31034 & \textbf{30886} & 30684 & \textbf{30663} (+)      \\
Li-30           & 34780 & \textbf{34606} & 34001 & \textbf{33745} (+)      \\
Li-31           & 36907 & \textbf{36599} & 36759 & \textbf{36748} \\
Li-32           & 39541 & \textbf{39134} & 39196 & \textbf{38696} (+)      \\
Loggi-n401-k23  & 40494 & \textbf{40423} & 40383 & {\ul 40417} (-)      \\
Loggi-n501-k24  & 18708 & \textbf{18515} (+)      & 18551 & {\ul 18665}    \\
Loggi-n601-k19  & 13402 & \textbf{13357} & 13195 & {\ul 13251}    \\
Loggi-n601-k42  & 34057 & \textbf{34005} & 33972 & {\ul 34007}    \\
Loggi-n901-k42  & 28828 & {\ul 28997}    & 28782 & {\ul 28732}    \\
Loggi-n1001-k31 & 26903 & {\ul 27032} (-)      & 26830 & {\ul 26867}    \\
ORTEC-n242-k12  & 17445 & {\ul 17466}          & 17461 & \textbf{17369} (+)      \\
ORTEC-n323-k21  & 23584 & \textbf{23490} & 23436 & \textbf{23376} \\
ORTEC-n405-k18  & 18345 & \textbf{18303} & 18260 & \textbf{18227} \\
ORTEC-n455-k41  & 19989 & \textbf{19963} & 19902 & {\ul 19939}    \\
ORTEC-n510-k23  & 30715 & \textbf{30629} (+)      & 30642 & {\ul 30692}    \\
ORTEC-n701-k64  & 41921 & \textbf{41895} & 41760 & {\ul 41844} (-)  \\
\bottomrule
\end{tabular}
\end{table}

\section{Conclusion}\label{sec:conclusion}
Search operators contribute to the major ability of metaheuristics. Each operator works by finding better solutions in the corresponding neighbourhood of a given solution. When the quality of the solution is better than all its neighbour solutions, i.e., the solution is a local optimum in the neighbourhood, further applying this operator brings no benefit. In this situation, we name the operator as a local optimum trapped operator. AOS approaches enhance the search efficiency of metaheuristics by dynamically and adaptively selecting operators during optimisation. However, due to the special characteristics of combinatorial optimisation problems, especially routing problems, the local optima of each operator significantly affects the performance of AOS approaches \cite{pei2022investigation}. Many resources are wasted in trying operators that have already been trapped in their local optima. 
To overcome the difficulty, the analysis of operators is considered an instructive method. Consistent characteristics of operators are expected to exist and able to help predict local optima.  
In this paper, we propose a novel method, named local optima correlation (LOC), that calculates the correlation between the local optima of the operators on a given problem. CVRP is taken as the test problem and various problem instances, including randomly generated instances and commonly used benchmark instances, are used to investigate the feature of LOC matrix. 
Experiment results on different sampling and instances indicate that the obtained LOC represents a general operator relationship. Then, an operator selection framework named AOS-assisted LOC is proposed, with the aim of predicting the local optima of each operator base on records from the early optimisation stage. By predicting the local optima, the resource wasted in applying local-optimum-trapped operators is reduced. The framework does not require any specific design of the base AOS approach. Therefore, a wide range of AOS approaches can be easily embedded. Experimental study and analysis verify the effectiveness of the proposed LOC-assisted AOS framework. 

As the difference of LOC from different instances can be evaluated, adaptively changing the effect degree of LOC on selection probability is a promising direction for further research. A larger range of instances and operators will be studied. Besides, LOC can be used to evaluate the similarity of problem instances from the view of operator and optimisation. It has a potential benefit on the wider research field besides operator selection.

% \begin{acks}
% This work was supported by the National Natural Science Foundation of China (Grant Nos. 62250710682, 61906083), the Guangdong Provincial Key Laboratory (Grant No. 2020B121201001), the Program for Guangdong Introducing Innovative and Enterpreneurial Teams (Grant No. 2017ZT07X386), the Shenzhen Science and Technology Program (Grant No. KQTD2016112514355531), the Shenzhen Fundamental Research Program (Grant No. JCYJ20190809121403553), and the Research Institute of Trustworthy Autonomous Systems.
% \end{acks}
\balance
\bibliographystyle{ACM-Reference-Format}
% \bibliography{sample-base, pricai2022aos}
\bibliography{camera_ready}


\begin{thebibliography}{29}

%%% the Web address.

\ifx \showCODEN    \undefined \def \showCODEN     #1{\unskip}     \fi
\ifx \showDOI      \undefined \def \showDOI       #1{#1}\fi
\ifx \showISBNx    \undefined \def \showISBNx     #1{\unskip}     \fi
\ifx \showISBNxiii \undefined \def \showISBNxiii  #1{\unskip}     \fi
\ifx \showISSN     \undefined \def \showISSN      #1{\unskip}     \fi
\ifx \showLCCN     \undefined \def \showLCCN      #1{\unskip}     \fi
\ifx \shownote     \undefined \def \shownote      #1{#1}          \fi
\ifx \showarticletitle \undefined \def \showarticletitle #1{#1}   \fi
\ifx \showURL      \undefined \def \showURL       {\relax}        \fi
% invisible to TeX
\providecommand\bibfield[2]{#2}
\providecommand\bibinfo[2]{#2}
\providecommand\natexlab[1]{#1}
\providecommand\showeprint[2][]{arXiv:#2}

\bibitem[\protect\citeauthoryear{Consoli and Yao}{Consoli and Yao}{2014}]%
        {consoli2014diversity}
\bibfield{author}{\bibinfo{person}{Pietro Consoli} {and} \bibinfo{person}{Xin
  Yao}.} \bibinfo{year}{2014}\natexlab{}.
\newblock \showarticletitle{Diversity-driven selection of multiple crossover
  operators for the capacitated arc routing problem}. In
  \bibinfo{booktitle}{\emph{Evolutionary Computation in Combinatorial
  Optimisation}}. \bibinfo{publisher}{Springer Berlin Heidelberg},
  \bibinfo{address}{Berlin, Heidelberg}, \bibinfo{pages}{97--108}.
\newblock
\showISBNx{978-3-662-44320-0}
\urldef\tempurl%
\url{https://doi.org/10.1007/978-3-662-44320-0_9}
\showDOI{\tempurl}


\bibitem[\protect\citeauthoryear{Consoli, Mei, Minku, and Yao}{Consoli
  et~al\mbox{.}}{2016}]%
        {consoli2016dynamic}
\bibfield{author}{\bibinfo{person}{Pietro~A. Consoli}, \bibinfo{person}{Yi
  Mei}, \bibinfo{person}{Leandro~L. Minku}, {and} \bibinfo{person}{Xin Yao}.}
  \bibinfo{year}{2016}\natexlab{}.
\newblock \showarticletitle{Dynamic selection of evolutionary operators based
  on online learning and fitness landscape analysis}.
\newblock \bibinfo{journal}{\emph{Soft Computing}} \bibinfo{volume}{20},
  \bibinfo{number}{10} (\bibinfo{date}{apr} \bibinfo{year}{2016}),
  \bibinfo{pages}{3889--3914}.
\newblock
\urldef\tempurl%
\url{https://doi.org/10.1007/s00500-016-2126-x}
\showDOI{\tempurl}


\bibitem[\protect\citeauthoryear{Croes}{Croes}{1958}]%
        {croes1958method}
\bibfield{author}{\bibinfo{person}{Georges~A Croes}.}
  \bibinfo{year}{1958}\natexlab{}.
\newblock \showarticletitle{A method for solving traveling-salesman problems}.
\newblock \bibinfo{journal}{\emph{Operations research}} \bibinfo{volume}{6},
  \bibinfo{number}{6} (\bibinfo{year}{1958}), \bibinfo{pages}{791--812}.
\newblock
\urldef\tempurl%
\url{https://doi.org/10.1287/opre.6.6.791}
\showDOI{\tempurl}


\bibitem[\protect\citeauthoryear{Durgut, Aydin, and Atli}{Durgut
  et~al\mbox{.}}{2021}]%
        {durgut2021adaptive}
\bibfield{author}{\bibinfo{person}{Rafet Durgut}, \bibinfo{person}{Mehmet~Emin
  Aydin}, {and} \bibinfo{person}{Ibrahim Atli}.}
  \bibinfo{year}{2021}\natexlab{}.
\newblock \showarticletitle{Adaptive operator selection with reinforcement
  learning}.
\newblock \bibinfo{journal}{\emph{Information Sciences}}  \bibinfo{volume}{581}
  (\bibinfo{year}{2021}), \bibinfo{pages}{773--790}.
\newblock
\showISSN{0020-0255}
\urldef\tempurl%
\url{https://doi.org/10.1016/j.ins.2021.10.025}
\showDOI{\tempurl}


\bibitem[\protect\citeauthoryear{Fialho, Da~Costa, Schoenauer, and
  Sebag}{Fialho et~al\mbox{.}}{2008}]%
        {fialho2008extreme}
\bibfield{author}{\bibinfo{person}{{\'A}lvaro Fialho},
  \bibinfo{person}{Lu{\'\i}s Da~Costa}, \bibinfo{person}{Marc Schoenauer},
  {and} \bibinfo{person}{Michele Sebag}.} \bibinfo{year}{2008}\natexlab{}.
\newblock \showarticletitle{Extreme value based adaptive operator selection}.
  In \bibinfo{booktitle}{\emph{Parallel Problem Solving from Nature -- PPSN
  X}}. \bibinfo{publisher}{Springer Berlin Heidelberg},
  \bibinfo{address}{Berlin, Heidelberg}, \bibinfo{pages}{175--184}.
\newblock
\showISBNx{978-3-540-87700-4}
\urldef\tempurl%
\url{https://doi.org/10.1007/978-3-540-87700-4_18}
\showDOI{\tempurl}


\bibitem[\protect\citeauthoryear{Fialho, Ros, Schoenauer, and Sebag}{Fialho
  et~al\mbox{.}}{2010}]%
        {fialho2010comparison}
\bibfield{author}{\bibinfo{person}{{\'A}lvaro Fialho}, \bibinfo{person}{Raymond
  Ros}, \bibinfo{person}{Marc Schoenauer}, {and} \bibinfo{person}{Mich{\`e}le
  Sebag}.} \bibinfo{year}{2010}\natexlab{}.
\newblock \showarticletitle{Comparison-Based Adaptive Strategy Selection with
  Bandits in Differential Evolution}. In \bibinfo{booktitle}{\emph{Parallel
  Problem Solving from Nature, PPSN XI}}. \bibinfo{publisher}{Springer Berlin
  Heidelberg}, \bibinfo{address}{Berlin, Heidelberg},
  \bibinfo{pages}{194--203}.
\newblock
\showISBNx{978-3-642-15844-5}
\urldef\tempurl%
\url{https://doi.org/10.1007/978-3-642-15844-5_20}
\showDOI{\tempurl}


\bibitem[\protect\citeauthoryear{Goldberg}{Goldberg}{1990}]%
        {goldberg1990probability}
\bibfield{author}{\bibinfo{person}{David~E Goldberg}.}
  \bibinfo{year}{1990}\natexlab{}.
\newblock \showarticletitle{Probability matching, the magnitude of
  reinforcement, and classifier system bidding}.
\newblock \bibinfo{journal}{\emph{Machine Learning}} \bibinfo{volume}{5},
  \bibinfo{number}{4} (\bibinfo{year}{1990}), \bibinfo{pages}{407--425}.
\newblock
\urldef\tempurl%
\url{https://doi.org/10.1023/A:1022681708029}
\showDOI{\tempurl}


\bibitem[\protect\citeauthoryear{Handoko, Nguyen, Yuan, and Lau}{Handoko
  et~al\mbox{.}}{2014}]%
        {handoko2014reinforcement}
\bibfield{author}{\bibinfo{person}{Stephanus~Daniel Handoko},
  \bibinfo{person}{Duc~Thien Nguyen}, \bibinfo{person}{Zhi Yuan}, {and}
  \bibinfo{person}{Hoong~Chuin Lau}.} \bibinfo{year}{2014}\natexlab{}.
\newblock \showarticletitle{Reinforcement learning for adaptive operator
  selection in memetic search applied to quadratic assignment problem}. In
  \bibinfo{booktitle}{\emph{Proceedings of the Companion Publication of the
  2014 Annual Conference on Genetic and Evolutionary Computation}} (Vancouver,
  BC, Canada) \emph{(\bibinfo{series}{GECCO Comp '14})}.
  \bibinfo{publisher}{Association for Computing Machinery},
  \bibinfo{address}{New York, NY, USA}, \bibinfo{pages}{193–194}.
\newblock
\showISBNx{9781450328814}
\urldef\tempurl%
\url{https://doi.org/10.1145/2598394.2598451}
\showDOI{\tempurl}


\bibitem[\protect\citeauthoryear{Li, Golden, and Wasil}{Li
  et~al\mbox{.}}{2005}]%
        {li2005very}
\bibfield{author}{\bibinfo{person}{Feiyue Li}, \bibinfo{person}{Bruce Golden},
  {and} \bibinfo{person}{Edward Wasil}.} \bibinfo{year}{2005}\natexlab{}.
\newblock \showarticletitle{Very large-scale vehicle routing: new test
  problems, algorithms, and results}.
\newblock \bibinfo{journal}{\emph{Computers \& Operations Research}}
  \bibinfo{volume}{32}, \bibinfo{number}{5} (\bibinfo{year}{2005}),
  \bibinfo{pages}{1165--1179}.
\newblock
\showISSN{0305-0548}
\urldef\tempurl%
\url{https://doi.org/10.1016/j.cor.2003.10.002}
\showDOI{\tempurl}


\bibitem[\protect\citeauthoryear{Lu, Zhang, and Yang}{Lu et~al\mbox{.}}{2020}]%
        {lu2020learning}
\bibfield{author}{\bibinfo{person}{Hao Lu}, \bibinfo{person}{Xingwen Zhang},
  {and} \bibinfo{person}{Shuang Yang}.} \bibinfo{year}{2020}\natexlab{}.
\newblock \showarticletitle{A learning-based iterative method for solving
  vehicle routing problems}. In \bibinfo{booktitle}{\emph{International
  Conference on Learning Representations}}.
\newblock
\urldef\tempurl%
\url{https://openreview.net/forum?id=BJe1334YDH}
\showURL{%
\tempurl}


\bibitem[\protect\citeauthoryear{Malan}{Malan}{2021}]%
        {malan2021survey}
\bibfield{author}{\bibinfo{person}{Katherine~Mary Malan}.}
  \bibinfo{year}{2021}\natexlab{}.
\newblock \showarticletitle{A survey of advances in landscape analysis for
  optimisation}.
\newblock \bibinfo{journal}{\emph{Algorithms}} \bibinfo{volume}{14},
  \bibinfo{number}{2} (\bibinfo{year}{2021}).
\newblock
\showISSN{1999-4893}
\urldef\tempurl%
\url{https://doi.org/10.3390/a14020040}
\showDOI{\tempurl}


\bibitem[\protect\citeauthoryear{Malan and Engelbrecht}{Malan and
  Engelbrecht}{2013}]%
        {malan2013survey}
\bibfield{author}{\bibinfo{person}{Katherine~M. Malan} {and}
  \bibinfo{person}{Andries~P. Engelbrecht}.} \bibinfo{year}{2013}\natexlab{}.
\newblock \showarticletitle{A survey of techniques for characterising fitness
  landscapes and some possible ways forward}.
\newblock \bibinfo{journal}{\emph{Information Sciences}}  \bibinfo{volume}{241}
  (\bibinfo{year}{2013}), \bibinfo{pages}{148--163}.
\newblock
\showISSN{0020-0255}
\urldef\tempurl%
\url{https://doi.org/10.1016/j.ins.2013.04.015}
\showDOI{\tempurl}


\bibitem[\protect\citeauthoryear{Maturana and Saubion}{Maturana and
  Saubion}{2008}]%
        {maturana2008compass}
\bibfield{author}{\bibinfo{person}{Jorge Maturana} {and}
  \bibinfo{person}{Fr{\'e}d{\'e}ric Saubion}.} \bibinfo{year}{2008}\natexlab{}.
\newblock \showarticletitle{A Compass to Guide Genetic Algorithms}. In
  \bibinfo{booktitle}{\emph{Parallel Problem Solving from Nature -- PPSN X}}.
  \bibinfo{publisher}{Springer Berlin Heidelberg}, \bibinfo{address}{Berlin,
  Heidelberg}, \bibinfo{pages}{256--265}.
\newblock
\showISBNx{978-3-540-87700-4}
\urldef\tempurl%
\url{https://doi.org/10.1007/978-3-540-87700-4_26}
\showDOI{\tempurl}


\bibitem[\protect\citeauthoryear{Mei, Tang, and Yao}{Mei et~al\mbox{.}}{2011}]%
        {mei2011decomposition}
\bibfield{author}{\bibinfo{person}{Yi Mei}, \bibinfo{person}{Ke Tang}, {and}
  \bibinfo{person}{Xin Yao}.} \bibinfo{year}{2011}\natexlab{}.
\newblock \showarticletitle{Decomposition-Based Memetic Algorithm for
  Multiobjective Capacitated Arc Routing Problem}.
\newblock \bibinfo{journal}{\emph{IEEE Transactions on Evolutionary
  Computation}} \bibinfo{volume}{15}, \bibinfo{number}{2}
  (\bibinfo{year}{2011}), \bibinfo{pages}{151--165}.
\newblock
\urldef\tempurl%
\url{https://doi.org/10.1109/TEVC.2010.2051446}
\showDOI{\tempurl}


\bibitem[\protect\citeauthoryear{Ochoa, Verel, Daolio, and Tomassini}{Ochoa
  et~al\mbox{.}}{2014}]%
        {ochoa2014local}
\bibfield{author}{\bibinfo{person}{Gabriela Ochoa},
  \bibinfo{person}{S{\'e}bastien Verel}, \bibinfo{person}{Fabio Daolio}, {and}
  \bibinfo{person}{Marco Tomassini}.} \bibinfo{year}{2014}\natexlab{}.
\newblock \bibinfo{booktitle}{\emph{Local Optima Networks: A New Model of
  Combinatorial Fitness Landscapes}}.
\newblock \bibinfo{publisher}{Springer Berlin Heidelberg},
  \bibinfo{address}{Berlin, Heidelberg}, \bibinfo{pages}{233--262}.
\newblock
\showISBNx{978-3-642-41888-4}
\urldef\tempurl%
\url{https://doi.org/10.1007/978-3-642-41888-4_9}
\showDOI{\tempurl}


\bibitem[\protect\citeauthoryear{Pei, Hu, Liu, Mei, and Yao}{Pei
  et~al\mbox{.}}{2021}]%
        {pei2021bi}
\bibfield{author}{\bibinfo{person}{Jiyuan Pei}, \bibinfo{person}{Chengpeng Hu},
  \bibinfo{person}{Jialin Liu}, \bibinfo{person}{Yi Mei}, {and}
  \bibinfo{person}{Xin Yao}.} \bibinfo{year}{2021}\natexlab{}.
\newblock \showarticletitle{Bi-objective splitting delivery VRP with loading
  constraints and restricted access}. In \bibinfo{booktitle}{\emph{2021 IEEE
  Symposium Series on Computational Intelligence (SSCI)}}.
  \bibinfo{pages}{01--09}.
\newblock
\urldef\tempurl%
\url{https://doi.org/10.1109/SSCI50451.2021.9659967}
\showDOI{\tempurl}


\bibitem[\protect\citeauthoryear{Pei, Mei, Liu, and Yao}{Pei
  et~al\mbox{.}}{2022}]%
        {pei2022investigation}
\bibfield{author}{\bibinfo{person}{Jiyuan Pei}, \bibinfo{person}{Yi Mei},
  \bibinfo{person}{Jialin Liu}, {and} \bibinfo{person}{Xin Yao}.}
  \bibinfo{year}{2022}\natexlab{}.
\newblock \showarticletitle{An investigation of adaptive operator selection
  in solving complex vehicle routing problem}. In
  \bibinfo{booktitle}{\emph{PRICAI 2022: Trends in Artificial Intelligence}}.
  \bibinfo{publisher}{Springer Nature Switzerland}, \bibinfo{address}{Cham},
  \bibinfo{pages}{562--573}.
\newblock
\showISBNx{978-3-031-20862-1}
\urldef\tempurl%
\url{https://doi.org/10.1007/978-3-031-20862-1_41}
\showDOI{\tempurl}


\bibitem[\protect\citeauthoryear{Sallam, Elsayed, Sarker, and Essam}{Sallam
  et~al\mbox{.}}{2017}]%
        {sallam2017landscape}
\bibfield{author}{\bibinfo{person}{Karam~M Sallam}, \bibinfo{person}{Saber~M
  Elsayed}, \bibinfo{person}{Ruhul~A Sarker}, {and} \bibinfo{person}{Daryl~L
  Essam}.} \bibinfo{year}{2017}\natexlab{}.
\newblock \showarticletitle{Landscape-based adaptive operator selection
  mechanism for differential evolution}.
\newblock \bibinfo{journal}{\emph{Information Sciences}}
  \bibinfo{volume}{418-419} (\bibinfo{year}{2017}), \bibinfo{pages}{383--404}.
\newblock
\showISSN{0020-0255}
\urldef\tempurl%
\url{https://doi.org/10.1016/j.ins.2017.08.028}
\showDOI{\tempurl}


\bibitem[\protect\citeauthoryear{Sharma, Komninos, L\'{o}pez-Ib\'{a}\~{n}ez,
  and Kazakov}{Sharma et~al\mbox{.}}{2019}]%
        {10.1145/3321707.3321813}
\bibfield{author}{\bibinfo{person}{Mudita Sharma}, \bibinfo{person}{Alexandros
  Komninos}, \bibinfo{person}{Manuel L\'{o}pez-Ib\'{a}\~{n}ez}, {and}
  \bibinfo{person}{Dimitar Kazakov}.} \bibinfo{year}{2019}\natexlab{}.
\newblock \showarticletitle{Deep Reinforcement Learning Based Parameter Control
  in Differential Evolution}. In \bibinfo{booktitle}{\emph{Proceedings of the
  Genetic and Evolutionary Computation Conference}} (Prague, Czech Republic)
  \emph{(\bibinfo{series}{GECCO '19})}. \bibinfo{publisher}{Association for
  Computing Machinery}, \bibinfo{address}{New York, NY, USA},
  \bibinfo{pages}{709–717}.
\newblock
\showISBNx{9781450361118}
\urldef\tempurl%
\url{https://doi.org/10.1145/3321707.3321813}
\showDOI{\tempurl}


\bibitem[\protect\citeauthoryear{Sharma, L{\'o}pez-Ib{\'a}{\~n}ez, and
  Kazakov}{Sharma et~al\mbox{.}}{2018}]%
        {sharma2018performance}
\bibfield{author}{\bibinfo{person}{Mudita Sharma}, \bibinfo{person}{Manuel
  L{\'o}pez-Ib{\'a}{\~n}ez}, {and} \bibinfo{person}{Dimitar Kazakov}.}
  \bibinfo{year}{2018}\natexlab{}.
\newblock \showarticletitle{Performance assessment of recursive probability
  matching for adaptive operator selection in differential evolution}. In
  \bibinfo{booktitle}{\emph{Parallel Problem Solving from Nature -- PPSN XV}}.
  \bibinfo{publisher}{Springer International Publishing},
  \bibinfo{address}{Cham}, \bibinfo{pages}{321--333}.
\newblock
\urldef\tempurl%
\url{https://doi.org/10.1007/978-3-319-99259-4_26}
\showDOI{\tempurl}


\bibitem[\protect\citeauthoryear{Soria~Alcaraz, Ochoa, Carpio, and
  Puga}{Soria~Alcaraz et~al\mbox{.}}{2014}]%
        {soria2014evolvability}
\bibfield{author}{\bibinfo{person}{Jorge~A Soria~Alcaraz},
  \bibinfo{person}{Gabriela Ochoa}, \bibinfo{person}{Martin Carpio}, {and}
  \bibinfo{person}{Hector Puga}.} \bibinfo{year}{2014}\natexlab{}.
\newblock \showarticletitle{Evolvability metrics in adaptive operator
  selection}. In \bibinfo{booktitle}{\emph{Proceedings of the 2014 Annual
  Conference on Genetic and Evolutionary Computation}} (Vancouver, BC, Canada)
  \emph{(\bibinfo{series}{GECCO '14})}. \bibinfo{publisher}{Association for
  Computing Machinery}, \bibinfo{address}{New York, NY, USA},
  \bibinfo{pages}{1327–1334}.
\newblock
\showISBNx{9781450326629}
\urldef\tempurl%
\url{https://doi.org/10.1145/2576768.2598220}
\showDOI{\tempurl}


\bibitem[\protect\citeauthoryear{Tan, Li, and Wang}{Tan et~al\mbox{.}}{2021}]%
        {tan2021differential}
\bibfield{author}{\bibinfo{person}{Zhiping Tan}, \bibinfo{person}{Kangshun Li},
  {and} \bibinfo{person}{Yi Wang}.} \bibinfo{year}{2021}\natexlab{}.
\newblock \showarticletitle{Differential evolution with adaptive mutation
  strategy based on fitness landscape analysis}.
\newblock \bibinfo{journal}{\emph{Information Sciences}}  \bibinfo{volume}{549}
  (\bibinfo{year}{2021}), \bibinfo{pages}{142--163}.
\newblock
\showISSN{0020-0255}
\urldef\tempurl%
\url{https://doi.org/10.1016/j.ins.2020.11.023}
\showDOI{\tempurl}


\bibitem[\protect\citeauthoryear{Tayarani-N. and Prügel-Bennett}{Tayarani-N.
  and Prügel-Bennett}{2014}]%
        {tayaraniN2014landscape}
\bibfield{author}{\bibinfo{person}{Mohammad-H. Tayarani-N.} {and}
  \bibinfo{person}{Adam Prügel-Bennett}.} \bibinfo{year}{2014}\natexlab{}.
\newblock \showarticletitle{On the Landscape of Combinatorial Optimization
  Problems}.
\newblock \bibinfo{journal}{\emph{IEEE Transactions on Evolutionary
  Computation}} \bibinfo{volume}{18}, \bibinfo{number}{3}
  (\bibinfo{year}{2014}), \bibinfo{pages}{420--434}.
\newblock
\urldef\tempurl%
\url{https://doi.org/10.1109/TEVC.2013.2281502}
\showDOI{\tempurl}


\bibitem[\protect\citeauthoryear{Thierens}{Thierens}{2005}]%
        {thierens2005adaptive}
\bibfield{author}{\bibinfo{person}{Dirk Thierens}.}
  \bibinfo{year}{2005}\natexlab{}.
\newblock \showarticletitle{An adaptive pursuit strategy for allocating
  operator probabilities}. In \bibinfo{booktitle}{\emph{Proceedings of the 7th
  Annual Conference on Genetic and Evolutionary Computation}} (Washington DC,
  USA) \emph{(\bibinfo{series}{GECCO '05})}. \bibinfo{publisher}{Association
  for Computing Machinery}, \bibinfo{address}{New York, NY, USA},
  \bibinfo{pages}{1539–1546}.
\newblock
\showISBNx{1595930108}
\urldef\tempurl%
\url{https://doi.org/10.1145/1068009.1068251}
\showDOI{\tempurl}


\bibitem[\protect\citeauthoryear{Tong, Minku, Menzel, Sendhoff, and Yao}{Tong
  et~al\mbox{.}}{2022}]%
        {tong2022what}
\bibfield{author}{\bibinfo{person}{Hao Tong}, \bibinfo{person}{Leandro~L.
  Minku}, \bibinfo{person}{Stefan Menzel}, \bibinfo{person}{Bernhard Sendhoff},
  {and} \bibinfo{person}{Xin Yao}.} \bibinfo{year}{2022}\natexlab{}.
\newblock \showarticletitle{What makes the dynamic capacitated arc routing
  problem hard to solve: insights from fitness landscape analysis}. In
  \bibinfo{booktitle}{\emph{Proceedings of the Genetic and Evolutionary
  Computation Conference}} (Boston, Massachusetts)
  \emph{(\bibinfo{series}{GECCO '22})}. \bibinfo{publisher}{Association for
  Computing Machinery}, \bibinfo{address}{New York, NY, USA},
  \bibinfo{pages}{305–313}.
\newblock
\showISBNx{9781450392372}
\urldef\tempurl%
\url{https://doi.org/10.1145/3512290.3528756}
\showDOI{\tempurl}


\bibitem[\protect\citeauthoryear{Toth and Vigo}{Toth and Vigo}{2002}]%
        {toth2002models}
\bibfield{author}{\bibinfo{person}{Paolo Toth} {and} \bibinfo{person}{Daniele
  Vigo}.} \bibinfo{year}{2002}\natexlab{}.
\newblock \showarticletitle{Models, relaxations and exact approaches for the
  capacitated vehicle routing problem}.
\newblock \bibinfo{journal}{\emph{Discrete Applied Mathematics}}
  \bibinfo{volume}{123}, \bibinfo{number}{1} (\bibinfo{year}{2002}),
  \bibinfo{pages}{487--512}.
\newblock
\showISSN{0166-218X}
\urldef\tempurl%
\url{https://doi.org/10.1016/S0166-218X(01)00351-1}
\showDOI{\tempurl}


\bibitem[\protect\citeauthoryear{Yafrani, Martins, Krari, Wagner, Delgado,
  Ahiod, and L\"{u}ders}{Yafrani et~al\mbox{.}}{2018}]%
        {yafrani2018fitness}
\bibfield{author}{\bibinfo{person}{Mohamed~El Yafrani},
  \bibinfo{person}{Marcella S.~R. Martins}, \bibinfo{person}{Mehdi~El Krari},
  \bibinfo{person}{Markus Wagner}, \bibinfo{person}{Myriam R. B.~S. Delgado},
  \bibinfo{person}{Belaïd Ahiod}, {and} \bibinfo{person}{Ricardo L\"{u}ders}.}
  \bibinfo{year}{2018}\natexlab{}.
\newblock \showarticletitle{A fitness landscape analysis of the travelling
  thief problem}. In \bibinfo{booktitle}{\emph{Proceedings of the Genetic and
  Evolutionary Computation Conference}}. \bibinfo{publisher}{{ACM}}.
\newblock
\urldef\tempurl%
\url{https://doi.org/10.1145/3205455.3205537}
\showDOI{\tempurl}


\bibitem[\protect\citeauthoryear{Zhang, Sun, and Xu}{Zhang
  et~al\mbox{.}}{2021}]%
        {zhang2021learning}
\bibfield{author}{\bibinfo{person}{Haotian Zhang}, \bibinfo{person}{Jianyong
  Sun}, {and} \bibinfo{person}{Zongben Xu}.} \bibinfo{year}{2021}\natexlab{}.
\newblock \showarticletitle{Learning to mutate for differential evolution}. In
  \bibinfo{booktitle}{\emph{2021 IEEE Congress on Evolutionary Computation
  (CEC)}}. \bibinfo{pages}{1--8}.
\newblock
\urldef\tempurl%
\url{https://doi.org/10.1109/CEC45853.2021.9504990}
\showDOI{\tempurl}


\bibitem[\protect\citeauthoryear{Zhang, Wu, Tao, Pei, Liu, and Yao}{Zhang
  et~al\mbox{.}}{2020}]%
        {zhang2020d}
\bibfield{author}{\bibinfo{person}{Qingquan Zhang}, \bibinfo{person}{Feng Wu},
  \bibinfo{person}{Yang Tao}, \bibinfo{person}{Jiyuan Pei},
  \bibinfo{person}{Jialin Liu}, {and} \bibinfo{person}{Xin Yao}.}
  \bibinfo{year}{2020}\natexlab{}.
\newblock \showarticletitle{{D-MAENS2}: a self-adaptive {D-MAENS} algorithm
  with better decision diversity}. In \bibinfo{booktitle}{\emph{2020 IEEE
  Symposium Series on Computational Intelligence (SSCI)}}.
  \bibinfo{pages}{2754--2761}.
\newblock
\urldef\tempurl%
\url{https://doi.org/10.1109/SSCI47803.2020.9308250}
\showDOI{\tempurl}


\end{thebibliography}

\end{document}